\documentclass[webpdf,contemporary,large]{oup-authoring-template}

\graphicspath{{Fig/}}

\theoremstyle{thmstyleone}%
\theoremstyle{thmstyletwo}%
\theoremstyle{thmstylethree}%

\renewcommand\thesection{\arabic{section}}
\renewcommand\thesubsection{\thesection.\arabic{subsection}}
\renewcommand\thesubsubsection{\thesubsection.\arabic{subsubsection}}
\makeatletter
\renewcommand{\@seccntformat}[1]{%
  \csname the#1\endcsname\quad 
}
\makeatother

\usepackage{titlesec}
\usepackage{placeins}
\titleformat{\section}
  {\sffamily\bfseries\Large\raggedright}
  {\arabic{section}}{0.2em}{}
\titleformat{\subsection}
  {\sffamily\bfseries\normalsize\raggedright}
  {\arabic{section}.\arabic{subsection}}{0.2em}{}
\makeatletter
\renewcommand{\@seccntformat}[1]{\csname the#1\endcsname\hspace{0.2em}}
\makeatother
\newcommand{\pkg}{FederatedRSF}

\begin{document}

\journaltitle{Journal Title Here}
\DOI{DOI added during production}
\copyrightyear{YEAR}
\pubyear{YEAR}
\vol{XX}
\issue{x}
\access{Published: Date added during production}
\appnotes{Paper}


\title[Short Article Title]{\pkg\ : Federated Random Survival Forests for Partially Overlapping Medical Data}

\author[1,2]{Maryam Moradpour\textsuperscript{\dag}\ORCID{0009-0001-0906-5842}}
\author[1,2]{Jonas Harriehausen\textsuperscript{\dag}\ORCID{0009-0002-9882-4616}}
\author[3]{Amirreza Aleyasin\textsuperscript{\dag}\ORCID{0000-0003-2742-7138}}
\author[3]{Lion Philipp Wolf}
\author[4]{Youngjun Park\ORCID{0000-0002-9963-7014}}
\author[1,2,3,$\ast$]{Anne-Christin Hauschild\ORCID{0000-0002-7499-4379}}

\address[1]{\orgname{Institute for Predictive Deep Learning in Medicine and Healthcare, Justus Liebig University Gießen, Germany}}
\address[2]{\orgname{Hessian Center for Artificial Intelligence (hessian.AI), Darmstadt, Germany}}
\address[3]{\orgname{Department of Medical Informatics, University Medical Center Göttingen, Germany}}
\address[4]{\orgname{Max Planck Institute for Biology of Ageing, Cologne, North Rhine-Westphalia, Germany}}

\corresp[$\ast$]{Corresponding author. 
\href{mailto:anne-christin.hauschild@uni-giessen.de}{anne-christin.hauschild@uni-giessen.de}\\
\textsuperscript{\dag}Maryam Moradpour, Jonas Harriehausen, and Amirreza Aleyasin contributed equally to this work.}

\received{Date}{0}{Year}
\revised{Date}{0}{Year}
\accepted{Date}{0}{Year}



\abstract{Multi-center survival prediction can improve robustness and generalizability, yet privacy regulations and institutional governance often prevent pooling patient-level clinical and genomic data across institutions. In practice, deployment is further complicated by feature-space heterogeneity, in which sites collect different covariates or use different sequencing panels, resulting in only partially overlapping feature sets. We present \pkg\, a Python package that implements federated random survival forests, aggregating locally trained survival trees and redistributing only feature-compatible trees to each site, enabling inference with partial overlap without sharing raw data. We evaluate \pkg\ on the GBSG2 breast cancer cohort distributed with the scikit-survival package, simulating feature heterogeneity across clients by withholding subsets of features, and assessing discrimination using Harrell's concordance index (C-Index) under repeated cross-validation and site-splits. The results demonstrated that the federated model can achieve performance comparable to that of the centralized training setting.}


\keywords[Abbreviations]{\pkg\, RSF, FL, MCCV, C-index}

\keywords[Availability and implementation:]{Availability and implementation: Source code is available at \url{https://github.com/HauschildLab/FederatedRSF}.
The package is distributed on PyPI as \texttt{federated-rsf} (\texttt{pip install federated-rsf}).}

\maketitle

\section{Introduction}
 
Accurate survival predictions are crucial in oncology research and clinical decision support, particularly for time-to-event outcomes that are frequently right-censored (i.e., the event has not occurred for some patients by the last follow-up), which require specialized modeling approaches \citep{ishwaran2008random}. Contemporary survival modeling increasingly uses various covariates, including clinical variables and molecular features obtained from sequencing assays. Flexible nonparametric models, such as random survival forests (RSF), have become widely adopted for the analysis of censored data \citep{ishwaran2008random}. However, single-center cohorts are often limited in both size and diversity, underscoring the need for multi-cohort analyses to improve robustness and generalizability \citep{kairouz2021advances}.
 
A significant challenge in pooling patient-level clinical and genomic data between different institutions is ensuring compliance with privacy regulations and institutional governance protocols, such as the EU General Data Protection Regulation (GDPR) and the U.S. HIPAA Privacy Rule \citep{gdpr2016, hipaa2000}. Federated learning (FL) effectively mitigates this challenge by enabling collaborative model development without centralizing raw data. Instead, it aggregates model updates or computed artifacts locally. As a result, FL has emerged as a widely adopted approach for leveraging privacy-sensitive distributed datasets \citep{mcmahan2017federated}.
 
Despite the promise of combining clinical covariates with sequencing-derived features for survival prediction, a significant deployment challenge is feature-space heterogeneity, in which participating sites do not use identical sets of covariates \citep{ye2023heterogeneous}. In targeted sequencing applications, different institutions often rely on various gene panels and assay designs. Comparative studies indicate that these panels may span diverse gene sets, necessitating analyses that focus on overlapping genes or that address panel mismatches \citep{quy2022interassay}. From the perspective of federated learning, these scenarios deviate from the standard horizontal FL assumption of a common feature space and instead pertain to vertical or heterogeneous FL frameworks that explicitly manage partial feature overlap \citep{yang2019federated}.
 
We introduce \pkg\, a Python package that implements federated random survival forests for decentralized survival modeling under partially overlapping covariate spaces. \pkg\ builds on Random Survival Forests for right-censored outcomes \citep{ishwaran2008random} and is released as an open-source Python package built on \texttt{scikit-survival}. We evaluate \pkg\ on the GBSG2 breast cancer survival dataset distributed with \texttt{scikit-survival}, simulating feature-space heterogeneity by withholding subsets of covariates across clients and comparing federated training to matched local and centralized reference models.

\section{Materials and methods}

\subsection{Input data}
We consider $K$ participating sites. Site $k$ holds a local covariate matrix $X_k$ and right-censored
survival outcomes $Y_k=(T_k,\delta_k)$, where $T_k$ denotes the observed event or censoring time and
$\delta_k\in\{0,1\}$ indicates whether the event occurred. The goal is collaborative survival prediction
across sites without transferring patient-level records.

\subsection{Federated random survival forests under partial feature overlap}
The proposed approach follows a local training and model federation scheme in which
each site trains a Random Survival Forest locally and subsequently exchanges
model trained survival trees. The federation step is designed to explicitly account for the feature space
heterogeneity by distributing only those survival trees that are usable under a site's local conditions and
feature availability.

In concrete terms, the method is carried out in three steps:
(1) Local model training. Each site trains a local RSF on its own data $(X_k,Y_k)$
using standard survival tree construction with censoring-aware splitting. The result is a
collection of survival trees (a local forest) representing model-level information derived
from local data.
(2) Tree aggregation and compatibility filtering. A central coordinator collects the
locally trained forests and forms a federated pool of candidate trees. For a target site $i$,
the coordinator determines which trees are compatible with the site's feature set by
checking whether all split features used in a tree are present in site $i$'s available covariates.
Every compatible tree is redistributed to site $i$, ensuring that inference remains well-defined
despite partial feature overlap across clients.
(3) Local inference and evaluation. Each site integrates the received compatible trees
into its local estimator set and performs prediction on local test data, yielding risk scores and
survival performance metrics.

No raw covariates, survival times, or individual-level outcomes are exchanged at any point.
communication is restricted to model objects (and optional lightweight metadata required to
ensure consistent feature naming). Full implementation details and pseudocode are provided in the Supplementary Methods.

\subsection{Experiments}
We evaluated \pkg\ on the GBSG2 breast cancer cohort distributed with \texttt{scikit-survival}
(\texttt{sksurv.datasets.load\_gbsg2}; $n=686$). The dataset contains eight clinical covariates and a
right-censored time-to-event outcome (event rate $\approx 44\%$).

To emulate feature-space heterogeneity across sites, we simulated $K=10$ clients by randomly withholding
35\% of the pre-one-hot covariates per client for each site split. Each client's data was one-hot encoded after feature dropping and aligned to the union of retained features across sites (missing features represented
as NaN). By combining the training sets for all clients in each site-split run, we obtain Centralized-SRF (Site Restricted Features) data splits.

Full details of the cross-validation protocol, compared configurations, and statistical testing are provided
in the Supplementary Methods.

\section{Implementation}
The package is written in Python and requires Python version 3.11 or newer. Throughout this paper, we refer to the method and software as \pkg\; the source repository is hosted as FRSF4POD on GitHub, and the package is distributed on PyPI as \texttt{federated-rsf} (\texttt{pip install federated-rsf}).
Furthermore, it requires the \texttt{scikit-survival} library ($\geq$0.27.0) for the base implementation of the Random Survival Forest.
This package implements federated random survival forests for partially overlapping data. It enables joint training of Random Survival Forest models across multiple clients without sharing raw data or discarding features unavailable to all clients.

\subsection{Installation.}
The package can be installed from PyPI using \texttt{pip install federated-rsf} or from source from GitHub using the installation instructions (\href{https://github.com/HauschildLab/FederatedRSF#user-installation}{link}).

\subsection{Data Formats.}
The main input format for this package is tabular data, with a pandas DataFrame for the X values and a structured NumPy array for the y values.
The y values should contain a boolean column 'censor' and a float column 'time'.
One intermediary data format is the DataSchema, which stores the feature names of a dataset and a mapping, as a dictionary, from local to federated feature names.

\subsection{Components.}
The main components of the package with regard to the input data and preprocessing are the \texttt{DataSchema}, \texttt{SchemaCreator}, and \texttt{SchemaAligner}.
The \texttt{DataSchema} is a dataclass that contains feature names and optionally a mapping from local to centralized feature names.
The \texttt{SchemaCreator} is used to create a global \texttt{DataSchema} from all local schemas.
The \texttt{SchemaAligner} is used to apply the \texttt{DataSchema} received from the \texttt{SchemaCreator} to the local dataset.

With regard to the model training itself, the main components are the \texttt{LocalRandomSurvivalForest} and \texttt{FederatedRandomSurvivalForest}.
The \texttt{LocalRandomSurvivalForest} is trained on the local data.
The \texttt{FederatedRandomSurvivalForest} is used to collect and redistribute the trees of all \texttt{LocalRandomSurvivalForests}.

\subsection{Workflow.}
\begin{figure*}[!t]
    \centering
    \includegraphics[width=\textwidth]{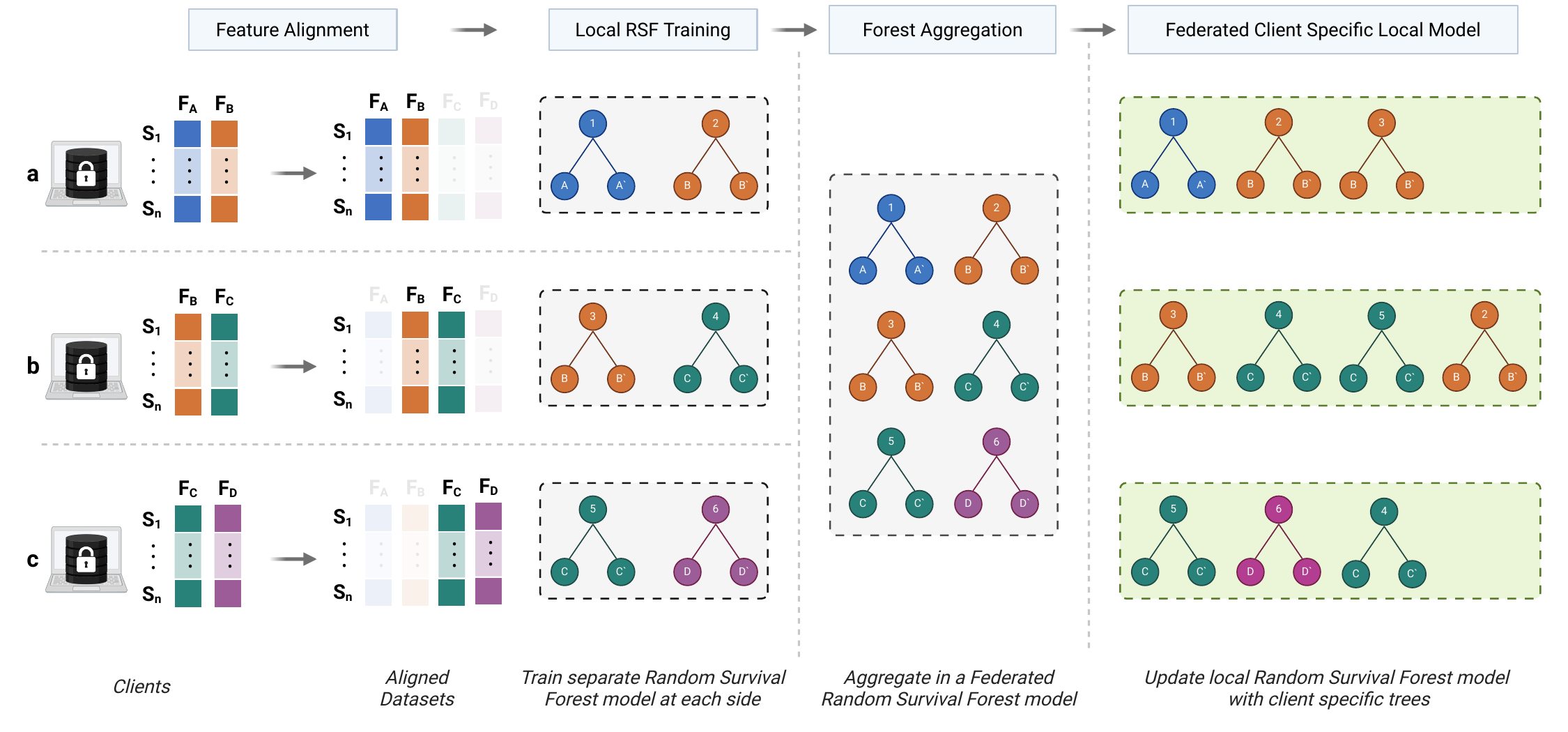}
    \caption{\textbf{Overview of the \pkg\ pipeline.}
    Each client holds a local dataset with partially overlapping features.
    The datasets are aligned to a common feature space by adding stub features in each local dataset.
    Each client then independently trains a local RSF model, producing a set of survival trees without sharing raw data.
    In the aggregation step, the locally trained trees are combined in a federated RSF model.
    Each client receives all trees fit for their local feature sets.
    Finally, the clients extend their local survival trees with the survival trees received from the federated RSF.
    This figure was created with BioRender.com.
    }
    \label{fig:workflow_overview}
\end{figure*}

For a graphical illustration of the Workflow, see Figure \ref{fig:workflow_overview}
\begin{itemize}
    \item Clients assign common names to all features with the same data, if they don't already have the same name.
    e.g. $AGE$, $age$, $age\_datetime$ $\xrightarrow{}$ $age$
    \item Clients create a DataSchema using their local feature names and the mapping from the previous step
    \item all DataSchemas are collected and aggregated to create a federated DataSchema with feature maps for each client
    \item Clients apply the federated DataSchema to their Dataset and train their local model.
    \item The local models are aggregated and receive all trees from other models that fit their local features, and the resulting trees are redistributed to the clients.
    \item Clients integrate the federated trees into their system according to their update strategy.
    \item Clients use the finished model.
\end{itemize}

\subsection{Reproducibility.}
All functions and classes utilizing randomness include a parameter $random\_state$ to allow for reproducible results.
The scripts to reproduce the figures in this paper and the supplementary material can be found in the examples folder in the GitHub repository (\href{https://github.com/HauschildLab/FederatedRSF/tree/main/examples}{link}).

\FloatBarrier

\section{Results}

\begin{figure*}[!t]
    \centering
    \includegraphics[width=0.65\textwidth]{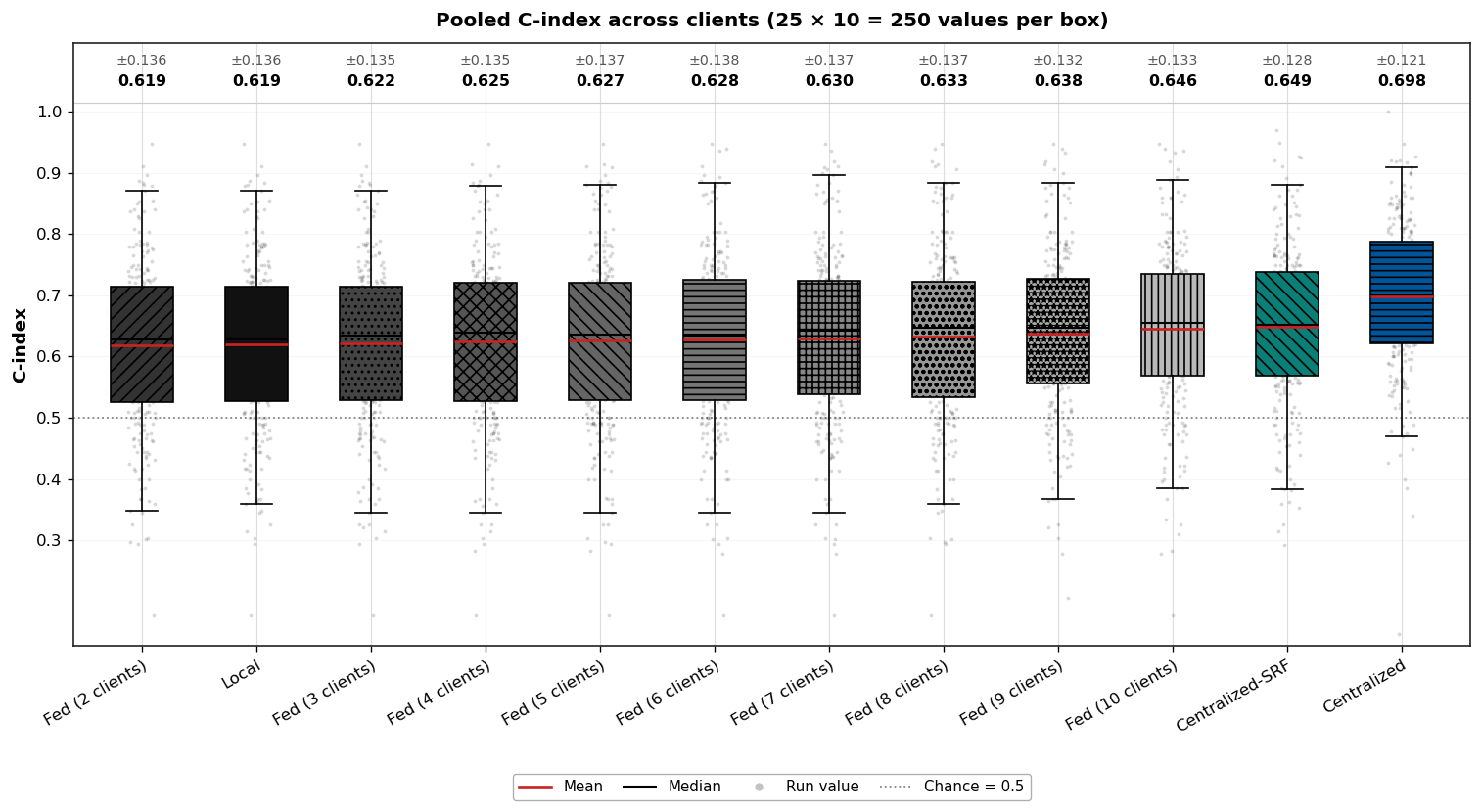}
    \caption{\textbf{Pooled C-index across configurations on the GBSG2
cohort.} Each box aggregates 250 evaluations
($5$ site splits $\times\,5$ folds $\times\,10$ client positions)
for one configuration on its own held-out test set. Numbers above
each box give the mean and standard deviation across runs.
Configurations are ordered by ascending mean C-index. \emph{Local}
denotes per-site RSFs without federation; \emph{Fed ($K$ clients)}
denotes \pkg\ at client count $K$; \emph{Centralized-SRF} denotes a
centralized RSF on the union of post-withholding features (no
privacy constraint, same restricted feature space as FederatedRSF);
\emph{Centralized} denotes an RSF on the original, unwithheld
cohort. The dotted line marks chance discrimination
($\mathrm{C}=0.5$).}
    \label{fig:cindex}
\end{figure*}

Pooled discrimination on each client's local test set is summarized
in Figure~\ref{fig:cindex}. Local training reached a mean C-index of
$0.619 \pm 0.136$. \pkg\ improved monotonically with the number of
participating clients, from $0.619 \pm 0.136$ at $K=2$ to
$0.646 \pm 0.133$ at $K=10$. The coincidence between $K=2$ and the
local baseline is mechanical: with only one peer per client, the
compatibility filter rarely admits cross-site trees beyond those each
client already trains locally, and federation gains accumulate only
from $K \ge 3$.

Paired comparisons across the 250 matched runs confirm that the gains
from federation are statistically significant and that \pkg\ recovers all of the discrimination available under the heterogeneous
covariate space. Moving from Local to \pkg\ at $K=10$ improved the
C-index by $+0.027$ on average (median $+0.018$; Wilcoxon signed-rank
$p = 1.9 \times 10^{-8}$; paired $t$-test $p = 2.2 \times 10^{-9}$).
Centralized-SRF reached a comparable mean of $0.649 \pm 0.128$, and the difference between Centralized-SRF and \pkg\ at $K=10$ was not statistically significant (mean $\Delta = +0.004$, median
$\Delta = 0.000$, Wilcoxon $p = 0.51$, paired $t$-test $p = 0.46$). For context, moving directly from Local to Centralized-SRF yielded $+0.030$ on average (Wilcoxon $p = 1.1 \times 10^{-6}$, paired $t$-test $p = 2.1 \times 10^{-7}$) — nearly identical to the gain achieved by \pkg\ without pooling raw data. \pkg\ is therefore statistically indistinguishable from a centralized model trained under the same feature constraint, while exchanging only the model artifacts.

The fully Centralized RSF set the empirical upper bound at
$0.698 \pm 0.121$. The $0.049$-point mean gap between Centralized
and Centralized-SRF quantifies the irreducible discrimination loss
imposed by panel heterogeneity itself, independent of training
scheme. Tree-level federation recovers, within statistical
uncertainty, the full performance attainable under the heterogeneous covariate space.

\section{Discussion and conclusion}\label{sec:discussion_conclusion}

\pkg\ supports federated survival prediction under privacy constraints and partially overlapping feature spaces by exchanging only feature-compatible survival trees. In the GBSG2 evaluation, federation improved over local training and was statistically indistinguishable from a centralized model trained under the same feature constraint; the residual gap to fully withheld training reflects feature heterogeneity itself. The method is provided as an open-source Python package with reproducible scripts and Supplementary Methods.

\section*{Conflicts of interest}
The authors declare that they have no competing interests.

\section*{Funding}
This work was supported by the Federal Ministry of Research, Technology, and Space (BMFTR) under grant agreements 01D2208A and 01KD2414A (FAIrPaCT), and by the hessian.AI Innovation Lab funded by the Hessian Ministry for Digital Strategy and Innovation (S-DIW04/0013/003).

\section*{Data availability}
The GBSG2 dataset is available via \texttt{scikit-survival}:
\url{https://scikit-survival.readthedocs.io/en/stable/api/datasets.html}. 
                             
Scripts to reproduce all analyses and figures are available at: \url{https://github.com/HauschildLab/FederatedRSF}.

\section*{Acknowledgments}
We acknowledge support from KISSKI (01IS22093A-E) and the DFG (grant no. 405797229). Computational resources were provided by GWDG and NHR-Nord@Göttingen (Emmy and Grete systems).


\bibliographystyle{oup-abbrvnat}
\bibliography{reference_FedRSF}

\end{document}


\maketitle

\section{Materials and Methods}\label{sec:supp_methods}

\subsection{Study design}\label{sec:study_design}
We evaluated \pkg\ in a multi-cohort survival prediction setting by treating each cohort as a federated client.
The overall workflow followed four stages: (i) dataset acquisition, (ii) preprocessing and feature harmonization,
(iii) model training under three analysis settings (local, federated, and pooled/global), and (iv) repeated validation
and statistical comparison. Predictive performance was estimated using repeated Monte-Carlo cross-validation (MCCV),
and model discrimination was quantified using Harrell's concordance index (C-index).

\subsection{Experimental setup (GBSG2 evaluation)}\label{sec:gbsg2_setup}
We evaluated FRSF on the GBSG2 breast cancer cohort distributed 
with \texttt{scikit-survival} 
(\path{sksurv.datasets.load_gbsg2}). The dataset includes $n=686$ node-positive primary breast cancer
patients and eight pre-one-hot clinical covariates (hormonal therapy, age, menopausal status, tumor size,
tumor grade, number of positive nodes, progesterone receptor, estrogen receptor). The outcome is
right-censored time-to-event; the overall event rate is $\approx 44\%$ ($299$ events), with the remainder censored.

Feature-space heterogeneity was simulated across $K=10$ clients by randomly withholding 35\% of the pre-one-hot
covariates per client for each site split (independent across clients and fixed across that split's folds). Each client's
data were one-hot encoded after feature dropping and aligned to the union of retained features across sites
(missing features represented as NaN).

We drew 5 random patient partition splits and ran 5 per-client cross-validation folds within each partition, yielding
$5\times 5 = 25$ runs per client and $25\times 10 = 250$ evaluations.

Four configurations were compared under a fixed RSF learner:
(i) \emph{Local}, a per-site RSF trained on the site's retained feature subset without federation;
(ii) \emph{FRSF} at varying client counts $k\in\{1,\ldots,10\}$;
(iii) \emph{Centralized-SRF} (site-restricted features), a single RSF trained without privacy constraints on the union of
post-withholding features, providing a feature-matched privacy-free reference for FRSF; and
(iv) \emph{Centralized}, a single RSF trained on the original cohort with all covariates retained, representing the empirical
upper bound in the absence of feature heterogeneity. All federated updates used the \texttt{constant} method, which samples
exactly $n_{\text{estimators}}$ trees from the combined local+federated pool (equal weighting). Paired differences across the
250 matched runs were tested using the Wilcoxon signed-rank test and a paired \textit{t}-test.

\subsection{Preprocessing and feature construction}\label{sec:preprocessing}
Preprocessing was performed per cohort using a shared pipeline to obtain machine-learning ready datasets.
First, patients without overall survival information or without mutation data were excluded. In addition, outlier samples
with extreme hypermutation patterns (e.g., POLE-driven ultra-hypermutation in TCGA) were removed to reduce noise
in mutation-derived features.

Clinical features were cleaned and harmonized by removing non-informative metadata fields, resolving duplicated
patient entries, and manually harmonizing semantically equivalent variables across cohorts (e.g., smoking status,
tumor purity scaling, metastasis indicators). To handle inconsistent nomenclature across studies, feature names were
standardized to a common vocabulary.

Mutation data were downloaded in mutation annotation format (MAF) and converted into a binary gene-by-patient matrix
indicating whether a gene was mutated in a patient. Multiple variants in the same gene were collapsed into a single
indicator. To reduce sparsity and likely passenger noise, genes mutated in only a single patient were removed. In
additional analyses, a knowledge-based filtering strategy was applied by restricting genes to cancer-relevant KEGG
pathways.

Finally, we defined both (i) \emph{local datasets} (one per cohort/client) and (ii) a \emph{pooled global dataset} as a
reference scenario where data sharing is allowed. Patient overlap across cohorts was handled by assigning duplicated
patients to a single cohort to preserve independence for federated/local analyses, whereas the pooled dataset aggregated
patient information to simulate centralized learning.

\subsection{Federated Random Survival Forest (FRSF) algorithm}\label{sec:frsf_algorithm}
\pkg\ extends Random Survival Forests (RSFs) to federated, feature-heterogeneous settings by exchanging survival trees
rather than raw records. Each client trains a local RSF and shares the trained trees with a coordinator. The coordinator
redistributes to each client only those trees whose split features are present in that client’s feature set
(\emph{tree compatibility}). This mechanism allows clients with different covariate panels to benefit from trees learned
elsewhere while preserving local feasibility and privacy.

\subsection{Model definitions}\label{sec:model_definitions}
We evaluated three model types:
(i) Local model: an RSF trained and evaluated using only a single site’s data.
(ii) Federated model: a site-specific RSF that is trained locally and then updated by incorporating compatible
trees received from other sites via \pkg\ federation.
(iii) Global model: an RSF trained on pooled data across all sites, representing a non-private upper-bound
reference where data sharing is permitted.

\subsection{Performance evaluation and statistical analysis}\label{sec:stat_analysis}
Model discrimination was assessed using Harrell’s concordance index (C-index), which accounts for right censoring and
evaluates whether predicted risk scores correctly rank survival times. To obtain robust performance estimates and
variability, we used 50 rounds of Monte-Carlo cross-validation (MCCV). In each round, data were randomly split into
training and test partitions (e.g., 70/30), and models were trained and evaluated independently. To ensure fair comparison,
identical splits (fixed random seed per run) were used when comparing local and federated models. Final performance was
reported as the mean C-index across MCCV runs with standard deviation.

When statistical comparisons were required, paired tests were performed across MCCV runs because the same split was
used for the compared models within each run. For non-inferiority comparisons between federated and global models, the
null hypothesis was defined as the federated model being worse than the global model by more than a pre-defined margin,
and significance was assessed at $\alpha=0.05$.

\section{Reproducibility}
\subsection{Environment Details}
The requirements of the package can be found in the GitHub Repository (\href{https://github.com/HauschildLab/FRSF4POD/blob/main/pyproject.toml}{link})
The exact versions of the relevant packages are:
\begin{itemize}
    \item Python 3.11.9
    \item scikit-learn              1.8.0
    \item scikit-survival           0.27.0
    \item pandas                    3.0.1
    \item numpy                     2.4.2
    \item joblib                    1.5.3
\end{itemize}

\subsection{Recreating Figures}

Install the package from pypi using
\begin{lstlisting}[language=bash]
pip install -U federated-rsf
\end{lstlisting}

Alternatively, install it from source by first cloning the GitHub repository and then running the following command at the repository root.
\begin{lstlisting}[language=bash]
pip install -U .
\end{lstlisting}

To reproduce the figures and tables in the paper run the scripts in the examples folder of the GitHub repository (\href{https://github.com/HauschildLab/FRSF4POD/tree/main/examples}{link})

\section{Full algorithm details}

\subsection{Step-by-Step federated RSF protocol}
The federated Random Survival Forest protocol proceeds in three main phases:
\emph{schema alignment}, \emph{model training}, and \emph{model federation}. The full workflow is as follows:

\begin{enumerate}
    \item \textbf{Feature name harmonisation.} Each client maps their local feature names
          to a common vocabulary (e.g.\ \texttt{AGE}, \texttt{age},
          \texttt{age\_datetime} $\rightarrow$ \texttt{age}).

    \item \textbf{Local schema creation.} Each client constructs a \texttt{DatasetSchema}
          from their local feature names and the map from local feature names to the common vocabulary.

    \item \textbf{Federated schema aggregation.} All local schemas are collected by a
          central coordinator and merged into a single \emph{federated schema} that
          contains per-client feature maps and covers the union of all features.

    \item \textbf{Schema alignment.} Each client applies the federated schema via a
          \texttt{SchemaAligner} to reindex their local dataset into the shared feature
          space.

    \item \textbf{Local model training.} Each client trains a local RSF on their
          aligned dataset.

    \item \textbf{Tree federation.} A \texttt{FederatedRandomSurvivalForest} aggregates
          the local models. Each client receives every tree from every other client
          whose feature set is compatible with that client's local features.

    \item \textbf{Model integration.} Each client integrates the federated trees
          according to their chosen \emph{update strategy} (\texttt{all} or
          \texttt{constant}).

    \item \textbf{Inference.} Clients use the finished federated model for prediction
          and evaluation.
\end{enumerate}

\subsection{Data Transfers}

\begin{itemize}
    \item \textbf{Upward (client $\to$ coordinator):} \texttt{DatasetSchema} objects
          containing local feature names and their canonical mappings. No raw data leaves the client at any point.

    \item \textbf{Downward (coordinator $\to$ clients):} The merged federated schema.

    \item \textbf{Upward (client $\to$ coordinator):} \texttt{LocalRandomSurvivalForest} objects
          containing the locally trained trees.

    \item \textbf{Downward (coordinator $\to$ clients):} The updated \texttt{LocalRandomSurvivalForest} object containing all trees that fit the local feature set.

    \item \textbf{Aggregation logic:} Tree compatibility is determined by feature
          overlap. A tree trained on client $j$ is sent to client $i$ if and only if
          every split feature used by that tree is present in client $i$'s aligned
          feature space. The receiving client's estimator list is then extended with
          these compatible remote trees.
\end{itemize}

\subsection*{Pseudocode}

\begin{algorithm}[H]
\caption{Federated Random Survival Forest Protocol}
\begin{algorithmic}[1]
\Require Clients $\{C_1, \dots, C_n\}$, each with local dataset $(X_i, Y_i)$

\medskip
\Statex \textit{--- Phase 1: Schema Alignment ---}

\For{each client $C_i$}
    \State Harmonise local feature names to canonical vocabulary
    \State Create feature map ${m}_i$ from local to canonical feature names
    \State $S_i \gets \texttt{DatasetSchema}(X_i.\text{columns}, {m}_i)$
    \State Send $S_i$ to coordinator
\EndFor

\State Coordinator: $\hat{S}_{\text{i}} \gets \texttt{SchemaCreator.fit\_transform}(\{S_1,\dots,S_n\})$
\For{each client $C_i$}
    \State Coordinator sends $\hat{S}_{\text{i}}$ to client $C_i$
\EndFor

\For{each client $C_i$}
    \State $\text{aligner}_i \gets \texttt{SchemaAligner.fit}(\hat{S}_{\text{i}})$
    \State $\hat{X}_i \gets \text{aligner}_i.\texttt{transform}(X_i)$
\EndFor

\medskip
\Statex \textit{--- Phase 2: Model Training ---}

\For{each client $C_i$ }
    \State Split $(\hat{X}_i, Y_i)$ into train / test sets
    \State $M_i \gets \texttt{LocalRSF.fit}(\hat{X}_i^{\text{train}},\, Y_i^{\text{train}})$
    \State Send model $M_i$ to coordinator
\EndFor

\medskip
\Statex \textit{--- Phase 3: Model Federation ---}

\State Coordinator: $M_{fed} \gets \bigcup_i M_i$
\State for-loop equivalent of $\hat{M}_i = M_{fed}.distribute\_trees()$:
\For{each client $C_i$}
    \State $\mathcal{T}_i^{\text{compat}} \gets \{t \in \mathcal{T} \setminus T_i \mid \text{features}(t) \subseteq \text{features}(\hat{X}_i)\}$
    \State Store $\mathcal{T}_i^{\text{compat}}$ on $M_i \to \hat{M_i}$
\EndFor
\State Send $M_i$ to $C_i$

\For{each client $C_i$}
    \State $M_i.\texttt{use\_federated\_estimators}()$
    \State results = $M_i.predict(X^{test}_i)$
\EndFor
\end{algorithmic}
\end{algorithm}

\subsection{Hyperparameters}

The tables below list every configurable parameter for each component of the
federated RSF framework, together with its default value, recommended range,
and the aspect of behaviour it controls.

\subsection*{DatasetSchema}

\begin{table}[H]
\centering
\caption{Parameters of \texttt{DatasetSchema}}
\begin{tabular}{L{6cm} L{1cm} L{3cm} L{5cm}}
\toprule
\textbf{Parameter} & \textbf{Default} & \textbf{Range / values} & \textbf{What it affects} \\
\midrule
\texttt{columns} & --- & list of str (required) &
    The local feature names used to construct the schema. \\[4pt]
\texttt{column\_map} & \texttt{None} & dict[str,str] or \texttt{None} &
    Maps local column names to canonical names. \texttt{None} signifies all local columns are the same as canonical names. \\[4pt]
\texttt{generate\_column\_map\_closure} & \texttt{True} & \texttt{True} / \texttt{False} &
    When \texttt{True}, extends \texttt{column\_map} with identity entries for any unmapped column, ensuring every feature has a canonical target. \\
\bottomrule
\end{tabular}
\end{table}

\subsection*{SchemaCreator}

\begin{table}[H]
\centering
\caption{Parameters of \texttt{SchemaCreator}}
\begin{tabular}{L{4cm} L{1.5cm} L{2cm} L{7.5cm}}
\toprule
\textbf{Parameter} & \textbf{Default} & \textbf{Range / values} & \textbf{What it affects} \\
\midrule
\texttt{anonymize} & \texttt{False} & \texttt{True} / \texttt{False} &
    Replaces canonical feature names with generic identifiers (\texttt{feature\_0}, \texttt{feature\_1}, \ldots), hiding semantics between clients. \\[4pt]
\texttt{extra\_columns} & \texttt{0} & $\geq 0$ integer &
    Reserves placeholder columns for clients joining after initial schema creation; set to at least the expected number of unseen features. \\[4pt]
\texttt{extra\_column\_prefix} & \texttt{"extra\_"} & any string &
    Prefix for placeholder column names; change if it conflicts with existing feature names. \\[4pt]
\texttt{random\_state} & \texttt{None} & integer or \texttt{None} &
    Seed for feature-name permutation when \texttt{anonymize=True}; fix for reproducibility. \\
\bottomrule
\end{tabular}
\end{table}

\subsection*{SchemaAligner}

\texttt{SchemaAligner} has no constructor parameters; it is configured
exclusively through its \texttt{fit} call.

\begin{table}[H]
\centering
\caption{Fit-time argument of \texttt{SchemaAligner}}
\begin{tabular}{L{3.5cm} L{1cm} L{3.5cm} L{7cm}}
\toprule
\textbf{Argument} & \textbf{Default} & \textbf{Range / values} & \textbf{What it affects} \\
\midrule
\texttt{dataset\_schema} & --- & \texttt{DatasetSchema} (required) &
    Supplies the full target schema and the column map used to rename and reindex the local DataFrame. Missing columns are filled with \texttt{NaN}. \\
\bottomrule
\end{tabular}
\end{table}

\subsection*{LocalRandomSurvivalForest}

Parameters inherited from \texttt{RandomSurvivalForest} (scikit-survival) are
included below alongside the federation-specific extensions.

\begin{table}[H]
\centering
\caption{Parameters of \texttt{LocalRandomSurvivalForest}}
\begin{tabular}{L{4cm} L{2cm} L{4.5cm} L{5cm}}
\toprule
\textbf{Parameter} & \textbf{Default} & \textbf{Range / values} & \textbf{What it affects} \\
\midrule
\multicolumn{4}{l}{\textit{Federation-specific parameters}} \\
\midrule
\texttt{update\_method} & \texttt{"all"} & \texttt{"all"} / \texttt{"constant"} &
    Controls how federated trees are integrated. \texttt{"all"} adds every compatible remote tree; \texttt{"constant"} samples trees to preserve the original \texttt{n\_estimators} count. \\[4pt]
\texttt{update\_weighting} & \texttt{"equal"} & \texttt{"equal"} / \texttt{"site\_size"} &
    Sampling weights used in \texttt{"constant"} mode. \texttt{"equal"} weights all trees uniformly; \texttt{"site\_size"} weights proportionally to the originating site's training set size. \\
\midrule
\multicolumn{4}{l}{\textit{Forest parameters (inherited from RandomSurvivalForest)}} \\
\midrule
\texttt{n\_estimators} & \texttt{100} & 50--500 &
    Number of trees; larger values improve stability at the cost of memory and compute. \\[4pt]
\texttt{max\_depth} & \texttt{None} & \texttt{None} or 1--20 &
    Maximum tree depth. \texttt{None} grows fully unpruned trees; restrict to reduce overfitting on small sites. \\[4pt]
\texttt{min\_samples\_split} & \texttt{6} & 2--20 &
    Minimum samples required to split an internal node; increase to smooth trees when site data are scarce. \\[4pt]
\texttt{min\_samples\_leaf} & \texttt{3} & 1--10 &
    Minimum samples required at a leaf; larger values regularise survival function estimates. \\[4pt]
\texttt{max\_features} & \texttt{"sqrt"} & \texttt{"sqrt"}, \texttt{"log2"}, int, float &
    Number of features considered at each split; \texttt{"sqrt"} is the standard default. \\[4pt]
\texttt{bootstrap} & \texttt{True} & \texttt{True} / \texttt{False} &
    Whether to use bootstrap sampling per tree. \texttt{False} uses the full local dataset for every tree. \\[4pt]
\texttt{oob\_score} & \texttt{False} & \texttt{True} / \texttt{False} &
    Enables out-of-bag C-index estimation without a separate validation set. \\[4pt]
\texttt{n\_jobs} & \texttt{None} & \texttt{None}, $-1$, or positive int &
    Parallelism for fit and predict; $-1$ uses all available cores. \\[4pt]
\texttt{random\_state} & \texttt{None} & integer or \texttt{None} &
    Seed for bootstrap and feature sampling; fix for reproducible experiments. \\[4pt]
\texttt{max\_samples} & \texttt{None} & \texttt{None}, int, or $(0,1]$ float &
    Samples drawn per tree when bootstrapping; reduce to speed up training on large sites. \\[4pt]
\texttt{low\_memory} & \texttt{False} & \texttt{True} / \texttt{False} &
    Reduces memory usage at predict time; disables \texttt{predict\_survival\_function} and \texttt{predict\_cumulative\_hazard\_function}. \\
\bottomrule
\end{tabular}
\end{table}

\subsection*{FederatedRandomSurvivalForest}

\texttt{FederatedRandomSurvivalForest} is a coordinator helper and cannot be
fit or used for prediction directly. Its only constructor parameter is listed below.

\begin{table}[H]
\centering
\caption{Parameters of \texttt{FederatedRandomSurvivalForest}}
\begin{tabular}{L{3cm} L{2.5cm} L{2.5cm} L{6cm}}
\toprule
\textbf{Parameter} & \textbf{Default} & \textbf{Range / values} & \textbf{What it affects} \\
\midrule
\texttt{local\_models} & --- & list of \texttt{LocalRSF} (required) &
    The fitted local models whose trees are aggregated and redistributed. All models must share the same \texttt{all\_features} ordering. \\
\bottomrule
\end{tabular}
\end{table}
